\documentclass[runningheads]{llncs}
\usepackage[T1]{fontenc}
\usepackage{graphicx}
\usepackage[bottom]{footmisc}
\usepackage{hyperref}
\usepackage{appendix}
\usepackage{xcolor}
\usepackage{longtable}
\usepackage{booktabs} 
\usepackage{multirow}

\usepackage[backend=biber,style=numeric,sorting=none]{biblatex} %Import biblatex package 
\begin{document}
\title{Is Mathematical Problem-Solving Expertise in Large Language Models Associated with Assessment Performance?\thanks{This manuscript has been accepted for presentation at the 28th International Conference on Human-Computer Interaction, to be held at the Montréal Convention Centre in Montréal, Canada, from 26 to 31 July 2026.}}
\author{Liang Zhang\inst{1}, Yu Fu\inst{2} \and Xinyi Jin \inst{3} 
} 
\authorrunning{L. Zhang et al.}
\institute{University of Michigan, Ann Arbor, MI, USA \\ \email{zhlian@umich.edu} \and
New York, NY, USA\\
\email{claireyufu@gmail.com} \and
The High School Affiliated to Minzu University of China, Beijing, China \\ \email{jinxichenzi@gmail.com}}
\maketitle              % typeset the header of the contribution
\begin{abstract}
Large Language Models (LLMs) are increasingly used in math education not only as problem solvers but also as assessors of learners’ reasoning. However, it remains unclear whether stronger math problem-solving ability is associated with stronger step-level assessment performance. This study examines that relationship using the GSM8K and MATH subsets of PROCESSBENCH, a human-annotated benchmark for identifying the earliest erroneous step in mathematical reasoning. We evaluate two LLM-based math tutor agent settings, instantiated with GPT-4 and GPT-5, in two independent tasks on the same math problems: solving the original problem and assessing a benchmark-provided solution by predicting the earliest erroneous step. Results show a consistent within-model pattern: assessment accuracy is substantially higher on math problem items the same model solved correctly than on items it solved incorrectly, with statistically significant associations across both models and datasets. At the same time, assessment remains more difficult than direct problem solving, especially on error-present solutions. These findings suggest that math problem-solving expertise supports stronger assessment performance, but reliable step-level diagnosis also requires additional capabilities such as step tracking, monitoring, and precise error localization. The results have implications for the design and evaluation of AI-supported Adaptive Instructional Systems (AISs) for formative assessment in math education.

\end{abstract}

\section{Introduction}
Large Language Models (LLMs) are increasingly being explored in math education for both instructional support and assessment \cite{pepin2025scoping,zhang2025mathematical}. Recent work suggests that LLMs are extending Adaptive Instructional Systems (AISs) beyond rule-based support toward richer dialogue, generative instructional assistance, and more flexible assessment workflows \cite{forsyth2024complex,zhang2024spl}. In math-related settings, LLMs have been studied for dialog-based tutoring \cite{gupta2025beyond}, hint generation \cite{qi2025tmath}, personalized feedback and error diagnosis \cite{reddig2025generating}, automated scoring of constructed responses \cite{morris2025automated}, proof grading \cite{zhao2025autograding}, and step-level verification of student reasoning \cite{daheim2024stepwise}. These developments position LLMs not only as math problem solvers, but also as potential assessors of students’ reasoning, making it important to understand how these two capabilities are related.

In human math education, assessing students’ reasoning depends not only on pedagogical skill, but also on sufficient understanding of the underlying math task and the diagnostic expertise needed to interpret students’ thinking \cite{shulman1986those,ball2008content}. We use this intuition as an analogy, rather than a direct equivalence, to motivate a similar question for LLMs: does stronger math problem-solving ability support stronger assessment of math reasoning? Drawing on Nelson and Narens’ distinction between object-level cognition and meta-level monitoring, we view problem solving as an object-level reasoning task, whereas identifying the earliest erroneous step in a provided solution is a meta-level monitoring task \cite{nelson1990metamemory}. From this perspective, the two capabilities should be related but not identical, because assessment requires more than solving alone, including tracking intermediate steps, checking consistency, and localizing where reasoning first breaks down. Although prior work has examined LLMs both as math problem solvers \cite{wei2022chain,wang2022self,chen2022program} and as assessment tools for grading student responses, identifying reasoning errors, and generating feedback or remediation \cite{caraeni2024evaluating}, these two capabilities have largely been studied separately.

Building on this distinction, we examine whether an LLM’s math problem-solving success is associated with its step-level assessment performance. This question is especially important for step-level assessment, because identifying the earliest error in a multi-step solution is not merely an outcome judgment; it requires understanding the original problem and determining where the reasoning first departs from a valid path. In educational settings, such diagnosis can support more interpretable and instructionally useful feedback than outcome-only scoring. Using the GSM8K and MATH subsets of PROCESSBENCH, a human-annotated benchmark for earliest-error identification in math reasoning \cite{zheng2025processbench}, we evaluate the same LLM-based math tutor agent on two independent tasks defined over the same underlying problems. In the problem-solving task, the tutor agent answers the original problem; in the assessment task, it identifies the earliest erroneous step in a benchmark-provided solution. More broadly, this study reframes LLM-based math assessment as a relationship between two related capabilities and informs the design of AI-supported assessment tools that can provide more interpretable and pedagogically useful feedback. 

\section{Methods}

\textbf{Dataset.} We use the GSM8K and MATH subsets of PROCESSBENCH \cite{zheng2025processbench} (\href{https://huggingface.co/datasets/Qwen/ProcessBench}{https://huggingface.co/datasets/Qwen/ProcessBench}), a human-annotated benchmark for identifying the earliest erroneous step in math reasoning. Each item includes an original math problem, a benchmark step-by-step solution trace, and a gold label indicating the earliest incorrect step using 0-based indexing, or $-1$ if the solution is fully correct. To create a balanced evaluation setting, we use all 400 GSM8K items and a randomly sampled 400-item subset of MATH, resulting in 800 evaluation problem items in total. See all processed data and experimental results at \href{https://github.com/LiangZhang2017/math-assessment-transfer}{https://github.com/LiangZhang2017/math-assessment-transfer}. 

\textbf{Models and task setup.} In both tasks, the system is instantiated as a single expert math tutor through a shared system prompt that frames the model as skilled in solving math problems and evaluating step-by-step solutions for errors. We evaluate two LLM-based math tutor configurations, using GPT-4 and GPT-5, on the same set of items in two independent tasks: problem solving and assessment. In the problem-solving task, the model receives only the original problem and is asked to generate a solution and final answer. In the assessment task, the model receives the same original problem together with the benchmark solution trace and is asked to identify the earliest erroneous step. For each configuration, the same model deployment is used across both tasks.  

\textbf{Evaluation and analysis.} Both tasks under each LLM model were repeated three times, and the mean results were reported to reduce run-to-run variability. Problem-solving performance was measured by final-answer accuracy, and assessment performance by exact-match accuracy on earliest-error prediction; following PROCESSBENCH, we also report assessment accuracy for error-present and no-error items and their harmonic mean (F1). We further compute a solve--assess gap as problem-solving accuracy minus assessment F1. To test the within-model association between solving and assessing, we compare assessment accuracy between solved-correct and solved-incorrect items using 2$\times$2 contingency tables with $\chi^2$ and Fisher’s exact tests, together with assessment-accuracy differences and 95\% confidence intervals. Finally, we conduct a brief qualitative analysis of representative agreement and mismatch cases to examine common failure modes and partial overlap between the two capabilities.  

\section{Results and Discussion}

\subsection{Overall Task Performance}

\begin{figure}[ht]
    \centering    \includegraphics[width=1\linewidth]{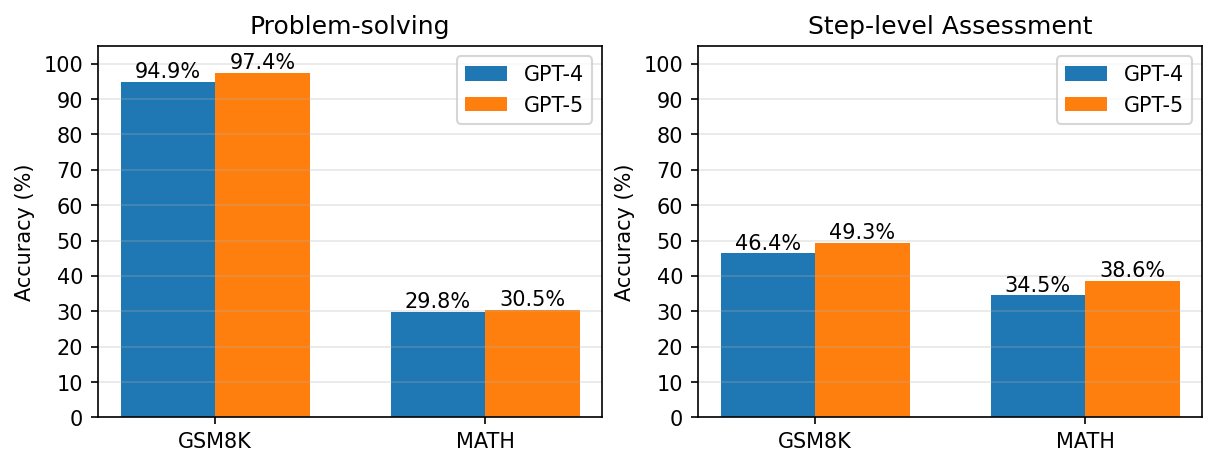}
    \caption{Accuracy on problem-solving (final answer vs. gold answer) and step-level assessment (predicted earliest error step vs. human label), by dataset and model (mean over 3 runs).}
    \label{fig:sovling_assessment}
\end{figure}

Fig.~\ref{fig:sovling_assessment} presents the performance of the LLM-based expert math tutor on the problem-solving task (left) and the step-level assessment task (right), averaged over three runs. On the problem-solving task, both models perform very strongly on GSM8K, with GPT-4 achieving 94.9\% accuracy and GPT-5 achieving 97.4\%, whereas performance drops sharply on MATH, where GPT-4 reaches 29.8\% and GPT-5 reaches 30.5\%. This contrast highlights the substantially greater difficulty of the MATH subset. On the step-level assessment task, performance is notably lower on GSM8K, with GPT-4 achieving 46.4\% and GPT-5 49.3\%, and remains moderate on MATH, where GPT-4 reaches 34.5\% and GPT-5 38.6\%. Across both tasks, GPT-5 consistently outperforms GPT-4, and this shared ordering is consistent with the hypothesis that stronger math problem-solving ability is associated with stronger assessment performance.

\subsection{Item-Level Association Between Problem-solving and Assessment}

\begin{figure}[ht]
    \centering
    \includegraphics[width=1\linewidth]{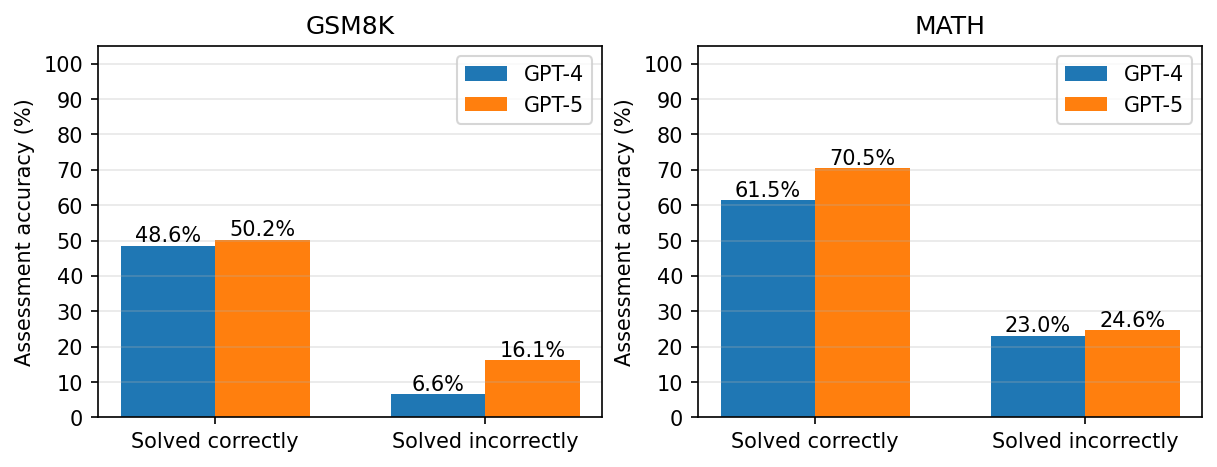}
    \caption{Step-level assessment accuracy grouped by whether the same LLM model solved the item correctly or incorrectly, by dataset and model (mean over 3 runs).}
    \label{fig:test}
\end{figure}

Fig.~\ref{fig:test} shows step-level assessment accuracy grouped by whether the same model solved the item correctly or incorrectly, averaged over three runs. A consistent pattern emerges across both models and both datasets: assessment accuracy is substantially higher on items the model solved correctly than on items it solved incorrectly. On GSM8K, GPT-4 achieves 48.6\% assessment accuracy on solved-correct items but only 6.6\% on solved-incorrect items, while GPT-5 reaches 50.2\% versus 16.1\%. A similar trend appears on MATH, where GPT-4 achieves 61.5\% on solved-correct items and 23.0\% on solved-incorrect items, and GPT-5 achieves 70.5\% versus 24.6\%. These results provide direct within-model evidence that successful problem solving on an item is associated with stronger step-level assessment on that same item. In other words, when the model is able to solve the problem correctly, it is much more likely to identify the earliest erroneous step in a benchmark solution.

We then tested the statistical significance of the association between solve outcome and assessment correctness using Fisher’s exact test on the corresponding 2$\times$2 contingency table, and report the difference in proportions with 95\% confidence intervals as an effect-size estimate. Table~\ref{tab:transfer-test} reports the within-model association between item-level solve outcome and step-level assessment accuracy. For each model and dataset, items were divided into solved-correct and solved-incorrect groups, and assessment accuracy was compared across these two groups. The $\chi^2$ statistics and Fisher’s exact test results both indicate a strong association between solve outcome and assessment correctness in every model--dataset condition. For GPT-4, the association was significant on both GSM8K ($\chi^2 = 39.38$, $p < 0.001$) and MATH ($\chi^2 = 162.32$, $p < 0.001$); for GPT-5, it was also significant on GSM8K ($\chi^2 = 12.71$, $p < 0.001$) and MATH ($\chi^2 = 224.33$, $p < 0.001$). The effect sizes, reported as the difference in assessment accuracy between solved-correct and solved-incorrect items, were consistently large, ranging from 34.1 to 45.9 percentage points. Specifically, GPT-4 showed differences of 42.0 points on GSM8K and 38.4 points on MATH, while GPT-5 showed differences of 34.1 points on GSM8K and 45.9 points on MATH. All 95\% confidence intervals excluded zero, indicating that these positive differences were robust across all settings. Taken together, these results show that, within the same LLM, successful problem solving on an item is strongly associated with better step-level assessment of that same item.
\begin{table*}[ht]
\centering
\scriptsize
\caption{Within-model association between item-level problem-solving success and step-level assessment accuracy. The table reports $\chi^2$ statistics, Fisher's exact test $p$ values, and differences in assessment accuracy (solved correct $-$ solved incorrect) with 95\% confidence intervals (CI), pooled over three runs.}
\label{tab:transfer-test}
\begin{tabular*}{\textwidth}{@{\extracolsep{\fill}}cccccc@{}}
\toprule
Model & Dataset & $\chi^2$ & $p$ (Fisher) & Diff (\%) & 95\% CI \\
\midrule
\multirow{2}{*}{GPT-4} 
& GSM8K & 39.38  & $5.35 \times 10^{-12}$ & 42.0 & [35.1, 48.9] \\
& MATH  & 162.32 & $1.82 \times 10^{-36}$ & 38.4 & [32.6, 44.2] \\

\multirow{2}{*}{GPT-5} 
& GSM8K & 12.71  & $1.66 \times 10^{-4}$  & 34.1 & [20.8, 47.4] \\
& MATH  & 224.33 & $8.28 \times 10^{-51}$ & 45.9 & [40.4, 51.4] \\
\bottomrule
\end{tabular*}
\end{table*}

\subsection{Error-Sensitive Assessment Performance}

Table~\ref{tab:processbench-metrics} reports step-level assessment metrics separated by whether the benchmark solution contains an error. Specifically, we report accuracy on error-present samples, accuracy on no-error samples, and their harmonic mean (F1), following the benchmark evaluation protocol \cite{zheng2025processbench}. Across both models and both datasets, accuracy is consistently high on no-error samples (72.9\%--97.1\%) but remains very low on error-present samples (4.7\%--10.0\%), resulting in low F1 values overall (8.9\%--17.5\%). GPT-5 achieves higher no-error accuracy than GPT-4 on both datasets, whereas error-present accuracy remains similarly limited across models and appears to be the main bottleneck. F1 is particularly informative here because it balances two different failure modes: incorrectly flagging errors in fully correct solutions and failing to detect the earliest erroneous step when an error is present. The results show that both models are much better at recognizing that a solution is fully correct than at localizing where a reasoning process first goes wrong. This helps explain why overall assessment accuracy can appear moderate while error-sensitive step-level assessment performance remains weak.

\begin{table*}[ht!]
\centering
\scriptsize
\caption{Step-level assessment metrics by error presence: accuracy on error-present samples (human label $\neq -1$), accuracy on no-error samples (human label $= -1$), and their harmonic mean (F1), pooled over three runs.}
\label{tab:processbench-metrics}
\begin{tabular*}{\textwidth}{@{\extracolsep{\fill}}ccccc@{}}
\toprule
Model & Dataset & Accuracy (error present) & Accuracy (no error) & F1 \\
\midrule
\multirow{2}{*}{GPT-4}
& GSM8K & 4.7  & 91.2 & 8.9 \\
& MATH  & 10.0 & 72.9 & 17.5 \\

\multirow{2}{*}{GPT-5}
& GSM8K & 4.8  & 97.1 & 9.2 \\
& MATH  & 7.4  & 87.4 & 13.6 \\
\bottomrule
\end{tabular*}
\end{table*}

\subsection{Assessment Difficulty Relative to Problem Solving}

Table~\ref{tab:solve-assess-gap} complements the baseline results in Fig.~\ref{fig:sovling_assessment} by summarizing two derived indicators of assessment difficulty: assessment F1 and the solve--assess gap. Assessment F1 captures step-level performance while balancing accuracy on error-present and no-error cases, and the gap shows how far assessment performance lags behind direct problem solving. On GSM8K, both models achieve very large gaps, with GPT-4 and GPT-5 reaching 86.0 and 88.2 percentage points, respectively, while assessment F1 remains below 10\% for both models. On MATH, the gaps are smaller but still substantial, at 12.3 points for GPT-4 and 16.9 points for GPT-5, with assessment F1 ranging from 13.6\% to 17.5\%. These results reinforce that step-level assessment is not simply solving in another form: even when a model performs well at producing correct final answers, its ability to detect and localize the earliest erroneous step remains much more limited.
\begin{table*}[t]
\centering
\scriptsize
\caption{Derived summary of step-level assessment difficulty, complementing Fig.~\ref{fig:sovling_assessment}. The table reports assessment F1 and the solve--assess gap, computed as solving accuracy minus assessment F1, pooled over three runs.}
\label{tab:solve-assess-gap}
\begin{tabular*}{\textwidth}{@{\extracolsep{\fill}}cccc@{}}
\toprule
Model & Dataset & Assess F1 (\%) & Gap (Solve $-$ F1) \\
\midrule
\multirow{2}{*}{GPT-4}
& GSM8K & 8.9  & 86.0 \\
& MATH  & 17.5 & 12.3 \\

\multirow{2}{*}{GPT-5}
& GSM8K & 9.2  & 88.2 \\
& MATH  & 13.6 & 16.9 \\
\bottomrule
\end{tabular*}
\end{table*}

\subsection{Qualitative Analysis of Divergent Cases}
The qualitative cases help explain why the relationship between problem solving and step-level assessment is strong but incomplete. We examine the four possible outcome combinations formed by solving correctness and assessment correctness, with particular attention to the two most informative patterns: \textbf{solve correct, assess incorrect} and \textbf{solve incorrect, assess correct}. To make these contrasts concrete, we present representative examples together with the full problem statement, the benchmark label, and the model's assessment rationale.

\textbf{Solve correct, assess incorrect.}
In these cases, the model produces the correct final answer when solving the problem itself but fails to identify the earliest erroneous step in the benchmark solution. That is, it recognizes that something is wrong, but mislocalizes where the reasoning first breaks down. For example:
\begin{itemize}
    \item \textbf{GSM8K item \texttt{(gsm8k-192)}.} The problem states: \textit{``Over 30 days, Christina had 12 good days initially; the first 24 days are given (8 good, 8 bad, 8 neutral); the next three days are good, neutral, good; how many good days were left in the month?''} The benchmark solution concludes that there are 0 good days left. The human annotator marks the \emph{first} error at step 2, where the solution states \textit{``the number of days left in the month is $30 - 24 = 6$ days''} and then builds the remainder of the reasoning on that mistaken structure. GPT-4-based math tutor instead predicted the first error at step 5. Its rationale was \textit{``Step 5 incorrectly states that the next three days add 3 more good days to the count, but only two of the next three days are good (good, neutral, good), so it should add 2, not 3.''} Thus, the model identified a genuine counting error, but attributed the first error to a later step rather than to the earlier point at which the solution first went off track. This example shows that solving the problem correctly does not guarantee precise localization of the earliest error in another solution.

    \item \textbf{MATH item \texttt{(math-509)}.} The problem states: \textit{``Compute $\tan 315^\circ$.''} The benchmark solution uses the reference angle $45^\circ$ and states that in the fourth quadrant \textit{``the sine is positive and the cosine is negative,''} then concludes that $\tan 315^\circ = -1$. The human label marks the first error at step 2, where this quadrant-sign claim appears. GPT-4-based math tutor predicted the first error at step 3 and gave the rationale \textit{``Step 3 contains a conceptual error: it incorrectly states that in the fourth quadrant, `the sine is positive and the cosine is negative.' In fact, in the fourth quadrant, sine is negative and cosine is positive.''} Here, the model identified the same conceptual mistake as the annotator, but assigned it to the wrong step index. The critique is therefore semantically correct but still fails the benchmark's exact-match criterion for earliest-error localization.
\end{itemize}

\textbf{Solve incorrect, assess correct.}
In these cases, the model fails to solve the original problem correctly, but still assesses the benchmark solution accurately, either by correctly predicting \texttt{-1} for a fully correct solution or by correctly identifying the earliest erroneous step. For example:
\begin{itemize}
    \item \textbf{GSM8K item \texttt{(gsm8k-238)}.} The problem states: \textit{``Charisma meditates 15 minutes twice per day and does 1 hour of yoga 5 days per week; how much time on meditation and yoga in 4 weeks?''} The benchmark solution computes $0.5$ hours/day of meditation $\times$ 7 days $\times$ 4 weeks $= 14$ hours, plus 5 hours/week of yoga $\times$ 4 weeks $= 20$ hours, for a total of 34 hours. All steps are correct, so the human label is \texttt{-1}. GPT-4-based math tutor's own solution to this problem was incorrect, but when asked to assess the benchmark solution, it correctly predicted \texttt{-1} and stated \textit{``All steps are correct. Each calculation accurately reflects the problem conditions, and the reasoning is valid throughout.''} This case shows that the model can recognize a valid reasoning process even when it cannot generate the correct solution on its own.

    \item \textbf{MATH item \texttt{(math-572)}.} The problem states: \textit{``Given $a+b=7$ and $a^3+b^3=42$, find $1/a + 1/b$ as a common fraction.''} The benchmark solution applies the sum-of-cubes identity, derives $ab = 43/3$, and computes $1/a + 1/b = (a+b)/(ab) = 7/(43/3) = 21/43$. The human label is \texttt{-1}, indicating that all steps are correct. GPT-4-based math tutor failed to solve this problem in the solving task, but when assessing the benchmark solution, it correctly predicted \texttt{-1} and gave the rationale \textit{``All steps are mathly correct: the sum of cubes identity is applied properly, the expressions are manipulated correctly, and the final answer is computed accurately using the given values.''} Again, the model's generative solution was incorrect while its step-level assessment was correct, suggesting that critique draws on capabilities that partially overlap with, but are not identical to, those required for solving.
\end{itemize}

\textbf{Summary.} These cases reveal two important asymmetries. First, solving correctly does not necessarily imply assessing correctly, as shown by \texttt{gsm8k-192} and \texttt{math-509}, where the model reached the correct final answer but mislocalized the earliest error in the benchmark solution. Second, solving incorrectly does not necessarily imply assessing incorrectly, as shown by \texttt{gsm8k-238} and \texttt{math-572}, where the model failed to solve the problem itself but still correctly judged the benchmark solution as error-free. Taken together, these examples support the claim that transfer from math problem solving to step-level assessment is real but incomplete, and that assessment relies in part on capabilities that are not fully redundant with generative solving. 

\section{Discussion}

Our results provide converging evidence that mathematical problem-solving expertise in an LLM-based math tutor is associated with stronger step-level assessment performance, although the relationship is conditional rather than deterministic. Across both models and datasets, assessment accuracy was consistently higher on items the same model solved correctly than on items it solved incorrectly. At the aggregate level, the GPT-5-based math tutor also outperformed the GPT-4-based math tutor on both problem-solving and step-level assessment tasks. These findings directly support the central question of this study and are consistent with the framing introduced in the paper: problem solving and step-level assessment rely on overlapping capabilities, but assessment additionally involves monitoring a provided reasoning process rather than generating an answer alone \cite{nelson1990metamemory}. At the same time, the relationship is clearly not one-to-one. Solving success increases the likelihood of accurate assessment, but does not guarantee it.

The results also show that step-level assessment remains substantially more difficult than problem solving, especially when the benchmark solution actually contains an error. The large solve--assess gaps, low F1 values, and much stronger performance on no-error than error-present items suggest that the core challenge is not simply deciding whether a solution looks acceptable, but identifying where the reasoning first becomes invalid. This distinction is educationally important. In practice, confirming that a solution is correct is less useful than diagnosing the step at which reasoning breaks down, because targeted feedback, remediation, and adaptive scaffolding depend on precise error localization rather than outcome judgment alone \cite{zheng2025processbench,daheim2024stepwise,morris2025automated}. Our findings therefore suggest that evaluations of LLM-based math assessment should prioritize error-sensitive, step-level metrics, rather than relying only on overall accuracy or final-answer agreement.

The qualitative mismatch cases further clarify why solving and assessment should be viewed as related but partially distinct capabilities. In some cases, the model solved the original problem correctly but still misidentified the earliest erroneous step in the benchmark trace, suggesting that generating a valid solution can be easier than monitoring another solution at fine granularity. In other cases, the model failed to solve the original problem but still correctly judged the benchmark solution, especially when the provided trace was coherent or fully correct. These asymmetries indicate that assessment draws on additional capabilities beyond answer generation, including step alignment, local consistency checking, and precise error localization. More broadly, the findings imply that stronger math solvers are not automatically reliable assessors, and that improving AI-supported math tutors may require targeted support for critique, verification, and process supervision in addition to stronger generation \cite{cobbe2021training,lightman2023let,zheng2025processbench}.

% \section{Limitations and Future Works}
Several limitations qualify these conclusions. We evaluate only two LLM deployments and two PROCESSBENCH subsets, and the benchmark consists of curated reasoning traces rather than authentic student work. The exact-match earliest-error metric is also strict and may undercount partially correct critiques that identify the correct issue but assign it to a nearby step. In addition, because solving and assessment were measured on the same items, the observed association should not be interpreted as a purely causal transfer effect, since shared item difficulty may influence both outcomes. Future work should test more models, use authentic student reasoning data, incorporate graded measures of critique quality, and examine interactive tutoring settings that integrate solving, assessment, feedback, and revision over time. 

\section{Conclusion}

This study examined whether math problem-solving expertise in LLM-based math tutor agents is associated with stronger step-level assessment performance. Using GPT-4 and GPT-5 on the GSM8K and MATH subsets of PROCESSBENCH, we found a consistent within-model pattern: assessment performance was substantially higher on items the same model solved correctly than on items it solved incorrectly. This suggests that solving and assessing rely on overlapping reasoning capabilities. At the same time, large solve--assess gaps, low performance on error-present solutions, and qualitative mismatch cases show that the two are not equivalent. Reliable earliest-error diagnosis requires more than generating a correct final answer; it also requires step tracking, consistency monitoring, and precise error localization. Overall, our findings show that stronger math problem-solving ability supports, but does not fully determine, step-level assessment performance. These results highlight the need to evaluate AI-supported math tutors not only as problem solvers, but also as reasoning-process assessors that can provide more interpretable and instructionally useful feedback. 
% \section{Acknowledgments}
% The authors thank all collaborators and colleagues who contributed feedback and discussion during the development of this work. 

\printbibliography
\end{document}